\documentclass[conference,a4paper]{IEEEtran}
\IEEEoverridecommandlockouts

\usepackage[cmex10]{amsmath} 
\usepackage{amssymb,amsfonts} 
\usepackage{bm,bbm}

\usepackage[numbers,sort&compress]{natbib}

\usepackage[hidelinks]{hyperref}
\usepackage{graphicx}
\graphicspath{{Figures/PDF/}{Figures/PNG/}}
\usepackage{dblfloatfix}
\usepackage{float}
\usepackage{booktabs}
\usepackage{siunitx}
\usepackage{multirow}
\usepackage{tabularx}

\usepackage{algorithm}
\usepackage{algorithmic}

\usepackage{paralist}
\usepackage{textcomp}
\usepackage{xcolor}
\usepackage{orcidlink}
\usepackage{texnames}

\usepackage{tikz}
\usetikzlibrary{shapes.geometric, arrows, positioning}

\newcommand{\mX}{\mathbf{X}}
\newcommand{\mM}{\mathbf{M}}

\newcommand{\mZ}{\mathbf{Z}}

\newtheorem{remark}{Remark}

\begin{document}

\title{
\Large
\uppercase{LDGuid: A Framework for Robust Change Detection via Latent Difference Guidance}
}

\author{	
    \IEEEauthorblockN{Jiaxuan Zhao}
	\IEEEauthorblockA{
        \textit{University of Toronto}\\
		\texttt{yoyoeric.zhao@alumni.utoronto.ca}
        }
	\and
    \IEEEauthorblockN{Ali Bereyhi}
	\IEEEauthorblockA{
        \textit{University of Toronto}\\
		\texttt{ali.bereyhi@utoronto.ca}
        }
    \thanks{The Official implementation of \texttt{LDGuid} is available at: \url{https://github.com/zjxyoyo/LDGuid}}
}

\maketitle
\begin{abstract}
Modern deep learning models for change detection (CD) often struggle to explicitly represent task-relevant semantic differences. This paper proposes the \textit{Latent Difference Guidance} (\texttt{LDGuid}) framework that explicitly learns and injects semantic differences into CD models. \texttt{LDGuid} deploys adversarial autoencoding to implement a difference embedding (DE) module. The DE module is pretrained via the information bottleneck method, restricting it to learn only task-relevant differences between pre- and post-event samples. The learned latent difference is then used as an explicit guidance signal in the CD model.~We validate \texttt{LDGuid} by integrating it into U-Net, BIT, and AERNet baselines for CD and evaluating it on LEVIR-CD, WHU-CD, SVCD, and CaBuAr datasets. Experimental results show that \texttt{LDGuid} enhances segmentation performance across all benchmarks, with particularly remarkable gains in challenging settings affected by spectral noise. The results further highlight the ability of \texttt{LDGuid} in incorporating domain knowledge, such as task-specific spectral indices. Our findings suggest that semantic difference learning can drastically enhance the robustness of CD in remote sensing. 


\end{abstract}


\section{Introduction}
\label{sec:intro}

Change detection (CD) in bitemporal satellite imagery is a fundamental task in remote sensing. It plays a critical role in various applications such as urban expansion monitoring \cite{UrbanExpansion}, disaster assessment \cite{disasterassessment}, and environmental surveillance \cite{D2A}. Due to their expressive power, deep learning models have become dominant components for CD \cite{CDBaseAI}.  These models learn from the data a mapping from a pair of co-registered images, i.e., \textit{pre-} and \textit{post-event} images, to a binary \textit{change map}. Among various proposals, sequence-to-sequence architectures with skip connections, e.g. U-Net \cite{ronneberger2015unetconvolutionalnetworksbiomedical}, and Transformer-based models such as the bitemporal image transformer (BIT) \cite{BIT} can achieve state-of-the-art performance due to their ability to efficiently encode long-range spatiotemporal correlations.

Although state-of-the-art CD models can efficiently learn the binary change map, a fundamental task remains challenging for them: determining effective \textit{semantic differences} with minimal background noise. This task arises in several applications where \textit{soft information} on scene changes is required rather than a change map, e.g., wildfire segmentation \cite{Cambrin_2023}. Naive approaches, such as feature concatenation \cite{FC-Siam} or internal difference enhancement \cite{STADE-CDNet}, often fail to fully capture the \textit{change of interest} from complex static backgrounds. This issue is starkly illustrated in a recent wildfire segmentation result on the California Burned Area (CaBuAr) dataset \cite{Cambrin_2023}, where training a U-Net over both pre- and post-fire images surprisingly degrades performance compared to a model that is trained on post-fire images alone. This result indicates that while the model is capable of learning binary change maps, it struggles to efficiently represent semantic differences. 

This work proposes a novel framework, named \textit{Latent Difference Guidance} (\texttt{LDGuid}), to explicitly extract \textit{semantic differences} from co-registered inputs and inject them into segmentation models. Instead of relying on the segmentation model to implicitly learn the changes, \texttt{LDGuid} employs a pre-trained module to learn a latent representation of semantic differences, which can be efficiently incorporated into the CD model. Experiments across various datasets demonstrate the superiority of \texttt{LDGuid} over state-of-the-art CD approaches. 

\subsection{Contributions}
In this paper, we propose \texttt{LDGuid} as a generic framework to enhance CD models by extracting semantic differences. Our main contributions are as follows:
\begin{inparaenum}
    \item[($i$)] We design a difference embedding (DE) module, which uses an autoencoder with an adversarial decoder to learn a latent representation of difference with minimal background noise. 
    \item[($ii$)] Invoking the information bottleneck method, we develop a pre-training loop that trains the DE module. The pre-trained DE is then utilized to guide the main CD model.
    \item[($iii$)] We demonstrate the versatility of \texttt{LDGuid} by deploying it into three benchmark architectures for CD, namely U-Net, BIT, and AERNet. 
    \item[($iv$)] We conduct comprehensive experiments on several urban building datasets, i.e., LEVIR-CD \cite{LEVIR}, WHU-CD \cite{WHU-CD}, and SVCD \cite{SVCD}, as well as the recent CaBuAr dataset for wildfire segmentation \cite{Cambrin_2023}. Our results depict that \texttt{LDGuid} consistently outperforms baseline approaches, particularly in complex scenarios with high spectral noise such as burn scar detection.
\end{inparaenum}

\subsection{Related Work}
\label{sec:related}

\paragraph*{Deep CD Models}
Due to natural spatial correlation in data, earlier deep learning architectures for CD rely on CNN-based sequence-to-sequence models such as U-Net \cite{Siamese,CD_Unet2}. These models often capture local correlations. To capture global dependencies, Transformers are widely used in the CD literature. Examples of Transformer-based CD models are BIT \cite{BIT} and SwinSUNet \cite{SwinSUNet}. Recent studies focus on hybrid architectures, which combine local and global dependencies in the data. For instance, AERNet \cite{AER} combines CNNs with edge refinement, and ChangeBind \cite{ChangeBind} integrates Siamese CNNs with a hybrid change encoder to capture multi-scale semantic changes. Other recent proposals include state space models (SSMs), e.g., ChangeMamba \cite{ChangeMamba}, which model spatio-temporal relationships with linear complexity.

\paragraph*{Semantic Difference Learning}
In various applications of CD, semantic differences, i.e., soft information on scene changes, are to be learned explicitly. Traditional approaches employ  \textit{difference enhancement modules (DEMs)}, e.g., WNet \cite{WNet} and TransUNetCD \cite{TransUNetCD}. These modules extract difference features by combining the latent representations of pre- and post-event samples. The semantic differences learned by DEMs often contain considerable background noise. Recent studies such as \cite{DDPMCD} and \cite{CMMAN} propose using generative models, e.g., denoising diffusion probabilistic models (DDPMs) and generative adversarial networks (GANs), to extract more robust features. Although these approaches can improve the performance of the end-to-end model, they fail to reduce the background noise in semantic differences. This follows from the fact that DEMs do not incorporate the target difference in their training objectives. Unlike these approaches, \texttt{LDGuid} explicitly learns a latent representation of semantic differences by directly incorporating it in the pre-training objective.

\section{Problem Setup}
We consider CD in a supervised setting. Let $(\mX, \tilde{\mX}) \in \mathcal{X} \times \mathcal{X}$ be a pair of co-registered samples, where $\mX$ and $\tilde{\mX}$ are the \textit{pre-event} and  \textit{post-event} samples, respectively. The goal is to design a model $F_{\theta}: \mathcal{X} \times \mathcal{X} \mapsto \mathcal{M}$, which predicts a mask $\mM \in \mathcal{M}$ that depends on semantic changes of the samples, e.g., detecting burned areas from pre-fire and post-fire images. The model $F_\theta$ is trained on a dataset $\mathcal{D} = \{(\mX_i, \tilde{\mX}_i, \mM_i)\}_{i=1}^n$.

\begin{remark}
  Note that $\mM$ does not necessarily include all changes. In wildfire segmentation, for instance, some differences between pre- and post-fire images are not fire-related and hence excluded from the mask.
\end{remark}

The main objective is to develop a framework that captures the \textit{related} differences between the samples $\mX$ and $\tilde{\mX}$ with minimal \textit{background noise}, i.e., minimal information on the differences between $\mX$ and $\tilde{\mX}$ that are not related to $\mM$.

\section{Methodology and Design}
\label{sec:method}
\begin{figure}[t]
    \vspace{-4mm}
    \centering
    \begin{tikzpicture}
    \node[anchor=south west,inner sep=0] (image) at (0,0) {\includegraphics[width=.7\linewidth]{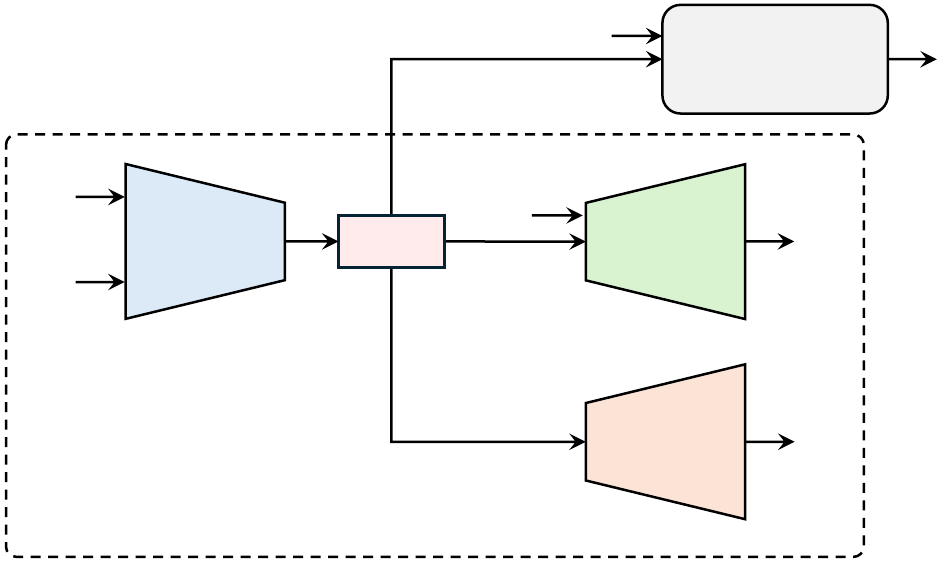}};
    \begin{scope}[x={(image.south east)},y={(image.north west)}]
        \node at (0.21,0.57) {$C_\phi$};
        \node at (0.05,0.65) {$\mX$};
        \node at (0.05,0.51) {$\tilde{\mX}$};
        \node at (0.415,0.57) {$\mZ$};
        \node at (0.53,0.63) {$\mX$};
        \node at (0.88,0.59) {$\hat{\mX}$};
        \node at (0.88,0.24) {$\Breve{\mX}$};
        \node at (0.72,0.57) { $D_\varphi$};
        \node at (0.72,0.22) { $A_\psi$};
        \node at (0.62,0.955) {$\tilde{\mX}$};
        \node at (0.83,0.9) {$G_\theta$};
        \node at (1.04,0.915) {$\hat{\mM}$};
        \node at (0.15,0.06) {\footnotesize{DE Module}};
    \end{scope}
    \end{tikzpicture}
    \caption{Overview of \texttt{LDGuid}: the dashed box shows the DE module. DE represents relevant semantic differences in a latent space.}\vspace{-.5cm}
    \label{fig:end-to-end}
    \vspace{-3mm}
\end{figure}
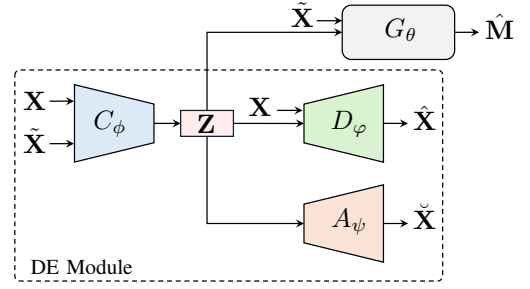

\texttt{LDGuid} is illustrated in Fig.~\ref{fig:end-to-end}. The framework consists of two components: ($i$) a difference embedding (DE) module that deploys adversarial autoencoding to learn a latent representation $\mZ$ for semantic differences, and ($ii$) a feature injection mechanism that incorporates $\mZ$ into the segmentation model.  

\subsection{Difference Embedding Module}
The DE module is shown in the dashed box in Fig.~\ref{fig:end-to-end}.~DE deploys adversarial autoencoding for semantic difference learning. Namely, the module consists of
\begin{enumerate}
    \item an autoencoder (AE) with encoder $C_\phi$ and decoder $D_\varphi$ that computes a latent representation $\mZ$. The decoder $D_\varphi$ estimates the post-event sample conditioned on the pre-event sample, i.e., it takes $\mZ$ and $\mX$ as input.  
    \item an \textit{adversarial} decoder $A_\psi$ that decodes the latent representation $\mZ$ \textit{unconditionally}, i.e., without having access to any other input rather than $\mZ$. 
\end{enumerate}
Note that $C_\phi$, $D_\varphi$, and $A_\psi$ are general learning models, e.g., deep neural networks (NNs), and are trained together. The DE module can hence be adapted to any data modality. Examples of these models for visual data are given in the experiments.


\paragraph*{Intuition on DE Design}
In the absence of the adversarial decoder $A_\psi$, the AE computes a latent representation~$\mZ$, which together with $\mX$ recovers $\hat{\mX} \approx \tilde{\mX}$. Intuitively, $\mZ$ contains semantic differences between $\mX$ and $\tilde{\mX}$. Nevertheless, it is not guaranteed that $\mZ$ does not contain background noise, i.e., irrelevant information. The role of adversarial decoder is to minimize background noise by decoding $\breve{\mX}$, which is highly different from $\tilde{\mX}$. This promotes $\mZ$ to contain \textit{only} relevant semantic differences required for recovery of $\tilde{\mX}$ from $\mX$ and suppress other irrelevant information in $\mX$ and $\tilde{\mX}$.


\subsection{Training DE Module}
Following the intuition on DE, we develop a mechanism based on the \textit{information bottleneck method} \cite{tishby2015deep} to train this module. To this end, let us formally define the features computed in this module: 
\begin{inparaenum}
\item[(1)] the encoder $C_\phi$ with learnable parameter $\phi$ computes the latent variable as $\mZ = C_\phi(\mX,\tilde{\mX})$,    
\item[(2)] the decoder $D_\varphi$ with learnable parameter $\varphi$ computes an estimate $\hat{\mX}$ of the post-event sample as $\hat{\mX} = D_\varphi (\mX,\mZ)$, and
\item[(3)] the adversarial decoder $A_\psi$ with learnable parameters $\psi$ computes the \textit{adversarial estimate} $\Breve{\mX}$ from the latent representation $\mZ$ as $\Breve{\mX} = A_\psi(\mZ)$.
\end{inparaenum}
As mentioned, DE aims to estimate the post-event sample $\tilde{\mX}$ at the output of the decoder, i.e., $\hat{\mX} \approx \tilde{\mX}$, while the adversarial estimate has minimal correlation with $\tilde{\mX}$. 
This can be obtained via the following loss.


\paragraph*{Training Loss}
Inspired by the information bottleneck method \cite{tishby2015deep}, we define a loss function by penalizing the AE recovery error with the fidelity of the adversarial estimate. Let $\mathcal{L}(\cdot): \mathcal{X} \times \mathcal{X} \mapsto [0,\infty)$ determine the loss between two data samples, e.g., mean squared error (MSE). We then define the loss of the DE module as
\begin{equation}
\mathcal{L}^{\mathrm{D}}_\beta (\mX, \tilde{\mX}) = \mathcal{L}(\hat{\mX}, \tilde{\mX}) - \beta \mathcal{L}(\Breve{\mX}, \tilde{\mX}),
\label{eq:autoencoder_loss}
\end{equation}
for some positive \textit{regularizer} $\beta$, which balances the trade-off between the quality of recovery by the AE and the level of adversarial suppression: while $\beta = 0$ trains a classical AE, $\beta\to\infty$ leads to a latent representation $\mZ$ that is independent of $\tilde{\mX}$. We treat $\beta$ as a hyperparameter.

The loss function $\mathcal{L}^{\mathrm{D}}_\beta (\cdot)$ implements the information bottleneck method. To see this point, let the risk be 
\begin{align}
{R}^{\mathrm{D}} (\phi,\varphi,\psi) = \mathbb{E} \left\lbrace \mathcal{L}^{\mathrm{D}}_\beta (\mX, \tilde{\mX}) \right\rbrace.
\end{align}
By minimizing this risk, we enforce $\hat{\mX}$ to be a valid estimate of the post-event sample $\tilde{\mX}$, and simultaneously the adversarial estimate $\Breve{\mX}$ to be minimally correlated with $\tilde{\mX}$. In this case, $\mZ$  carries \textit{only} semantic differences between $\mX$ and $\tilde{\mX}$, as it is the minimal information required by the decoder $D_\varphi(\cdot)$ to estimate $\tilde{\mX}$ from $\mX$ and $\mZ$. In practice, we estimate the risk ${R}^{\mathrm{D}} (\phi,\varphi,\psi)$ by averaging its samples computed from the dataset $\mathcal{D}$. 

\subsection{Latent Difference Injection}
Once the latent difference $\mZ$ is computed, it serves as an explicit guidance signal for the downstream CD task. As shown in Fig.~\ref{fig:end-to-end}, we inject $\mZ$ into the main segmentation model $G_\theta$ by treating it as input features. Depending on the architecture, $\mZ$ is resized and concatenated with the post-event image features or the backbone's feature maps. We discuss this procedure for U-Net, BIT, and AERNet in Section~\ref{sec:exp}. 

\subsection{Pretraining and Fine-tuning}
The DE module is initially pretrained on the dataset using the loss $\mathcal{L}^{\mathrm{D}}_\beta (\cdot)$. The pretrained DE is then kept frozen throughout the training of the main model. Alternatively, the DE can be fine-tuned by incorporating its loss into the main training loop. In this case, the total loss combines the segmentation loss $\mathcal{L}^{\mathrm{seg}}$, e.g., cross-entropy, and the DE loss as
\begin{equation}
\mathcal{L}^{\mathrm{total}} = \mathcal{L}^{\mathrm{seg}} (\mM, \hat{\mM}) + \lambda \mathcal{L}^{\mathrm{D}}_\beta (\mX, \tilde{\mX}),
\label{eq:total_loss}
\end{equation}
for some hyperparameter $\lambda$. 

The fine-tuning enables better suppression of background noise in the latent difference, as it adapts the latent variable to the underlying segmentation task. To stabilize training, we deploy a two-tier optimization strategy for fine-tuning in which the DE module is updated less frequently than the main model. Since the pre-trained DE module solely extracts features during CD, LDGuid's added computational overhead (both parameters and inference time) is negligible. 

\section{Experiments}
\label{sec:exp}
We next validate \texttt{LDGuid} by adapting it to different CD models and evaluating it on several CD datasets.  

\subsection{Experimental Setup}
\paragraph*{Datasets} 
We evaluate \texttt{LDGuid} on four remote sensing datasets. The first three are LEVIR-CD \cite{LEVIR}, WHU-CD \cite{WHU-CD}, and SVCD \cite{SVCD}, which capture structural and scene-level changes in urban environments. For wildfire application, we use CaBuAr \cite{Cambrin_2023}, which consists of multi-spectral Sentinel-2 imagery of burn scars. Furthermore, consistent gains across these datasets (e.g., LEVIR-CD, CaBuAr) prove \texttt{LDGuid}'s efficiency and scalability for large-region applications.

\paragraph*{Baselines and Metrics} We compare \texttt{LDGuid} against three baselines: ($i$) U-Net \cite{ronneberger2015unetconvolutionalnetworksbiomedical}, ($ii$) the Transformer-based architecture BIT \cite{BIT}, and ($iii$) the hybrid architecture AERNet \cite{AER}.
Unless otherwise stated, all models are trained for 200 epochs on an NVIDIA H100 GPU using Adam optimizer with learning rate $1 \times 10^{-4}$ and cross-entropy loss. Performance is evaluated using $F_1$-score and Intersection over Union (IoU).

\begin{figure*}[t]
    \vspace{-4mm}
    \centering
    \includegraphics[width=.7\linewidth]{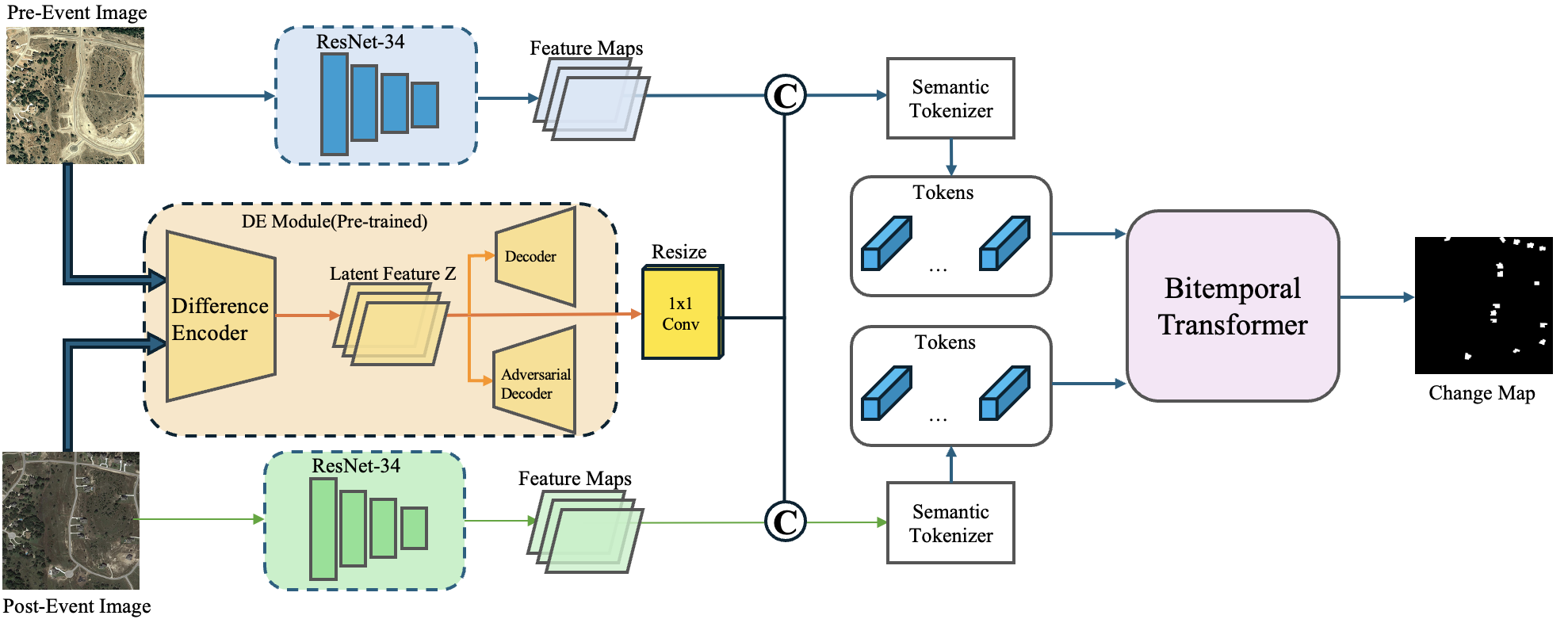}
    \caption{\texttt{LDGuid} with BIT architecture: latent difference is injected by concatenating the resized latent representation to the input of the semantic tokenizers.}
    \label{fig:injections}
    \vspace{-6mm}
\end{figure*}

\paragraph*{DE and Feature Injection}
\texttt{LDGuid} is adapted to all three baselines. As the DE module, we pre-train an adversarial AE considering a dual-branch encoder with two-layer CNNs to separately extract change and context latent features. These features are fused and fed into a three-layer transposed convolutional decoder, which computes both reconstruction and adversarial output. The latent difference $\mZ$ computed by the pre-trained DE is then injected into each baseline architecture as follows:
($i$) for {U-Net}, $\mZ$ is concatenated with the input post-event image.
($ii$) for {BIT} and {AERNet}, $\mZ$ is aligned and injected into the bottleneck, i.e., the features computed before the Transformer or GCFAM module. An example of \texttt{LDGuid} adaptation to BIT is shown in Fig.~\ref{fig:injections}.  

\subsection{Results and Discussion}
Table \ref{tab:main_results} summarizes the test results for \texttt{LDGuid} and baselines. The results are averaged over multiple seeds, with statistical significance assessed at $p < 0.05$. As observed, our framework yields consistent performance improvements across all datasets. Remarkably, on WHU-CD, \texttt{LDGuid} boosts the U-Net's IoU by {$23.36\%$} over the baseline. It further improves the IoU of the state-of-the-art BIT architecture by {$9\%$}, reaching $97.40\%$. The results confirm that the explicitly-learned semantic differences by DE compensate for the lack of global context in baseline architectures. Among the three baselines, AERNet depicts minimal gains. This is due to its internal edge refinement, which provides implicit information on semantic differences. Nevertheless, \texttt{LDGuid} still enhances performance in this case, which indicates that our latent difference guidance still provides complementary semantic information that cannot be fully captured from raw inputs alone.

\begin{table}[h!]
    \renewcommand{\arraystretch}{0.97}
    \vspace{-2mm}
    \centering
    \small
    \setlength{\tabcolsep}{2pt} 
    
    \caption{\normalfont IoU and $F_1$-score (\texttt{mean} $\pm$ \texttt{std}). Results reported in bold indicate best mean, and $^*$ denote statistical significance with $p < 0.05$.}
    \label{tab:main_results}
    
    \begin{tabular}{llcc}
        \toprule
        \textbf{Dataset} & \textbf{Method} & \textbf{IoU (\%)} & \textbf{$F_1$ (\%)} \\
        \midrule
        \multirow{6}{*}{\textbf{SVCD}} 
        & U-Net (Base) & $77.46 \pm 0.64$ & $87.09 \pm 0.41$ \\
        & \texttt{LDGuid} U-Net & $\mathbf{89.29 \pm 0.13}^*$ & $\mathbf{94.29 \pm 0.07}^*$ \\
        \cmidrule(lr){2-4}
        & BIT (Base) & $95.12 \pm 0.14$ & $97.46 \pm 0.08$ \\
        & \texttt{LDGuid} BIT & $\mathbf{96.08 \pm 0.02}^*$ & $\mathbf{97.98 \pm 0.01}^*$ \\
        \cmidrule(lr){2-4}
        & AERNet (Base) & $93.59 \pm 0.09$ & $96.67 \pm 0.05$ \\
        & \texttt{LDGuid} AERNet & $93.59 \pm 0.12$ & $96.67 \pm 0.05$ \\
        
        \midrule
        \multirow{6}{*}{\textbf{WHU}} 
        & U-Net (Base) & $58.37 \pm 0.72$ & $71.82 \pm 0.59$ \\
        & \texttt{LDGuid} U-Net & $\mathbf{81.73 \pm 1.20}^*$ & $\mathbf{89.44 \pm 0.75}^*$ \\
        \cmidrule(lr){2-4}
        & BIT (Base) & $88.31 \pm 1.48$ & $93.56 \pm 0.88$ \\
        & \texttt{LDGuid} BIT & $\mathbf{97.40 \pm 0.08}^*$ & $\mathbf{98.67 \pm 0.04}^*$ \\
        \cmidrule(lr){2-4}
        & AERNet (Base) & $81.43 \pm 1.33$ & $89.24 \pm 0.87$ \\
        & \texttt{LDGuid} AERNet & $\mathbf{82.50 \pm 1.13}$ & $\mathbf{89.87 \pm 0.74}$ \\
        
        \midrule
        \multirow{6}{*}{\textbf{LEVIR}} 
        & U-Net (Base) & $66.50 \pm 0.98$ & $79.83 \pm 0.70$ \\
        & \texttt{LDGuid} U-Net & $\mathbf{77.91 \pm 0.52}^*$ & $\mathbf{87.55 \pm 0.33}^*$ \\
        \cmidrule(lr){2-4}
        & BIT (Base) & $78.37 \pm 0.18$ & $86.53 \pm 0.14$ \\
        & \texttt{LDGuid} BIT & $\mathbf{83.19 \pm 0.12}^*$ & $\mathbf{90.07 \pm 0.09}^*$ \\
        \cmidrule(lr){2-4}
        & AERNet (Base) & $\mathbf{74.22 \pm 0.86}$ & $\mathbf{85.17 \pm 0.57}$ \\
        & \texttt{LDGuid} AERNet & $73.88 \pm 0.75$ & $84.94 \pm 0.51$ \\
        
        \midrule
        \multirow{8}{*}{\textbf{CaBuAr}} & \multicolumn{3}{l}{\textit{U-Net Baselines}} \\
        & \hspace{0.3em}- Baseline \cite{Cambrin_2023} (Post) & $58.30$ & $70.70$ \\
        & \hspace{0.3em}- Baseline \cite{Cambrin_2023} (Full) & $49.40$ & $62.50$ \\
        \cmidrule(lr){2-4}
        & \multicolumn{3}{l}{\textit{Baseline + Augmentation}} \\
        & \hspace{0.3em}- Post + Aug. & $75.36$ & $83.55$ \\
        & \hspace{0.3em}- Full + Aug. & $74.11$ & $83.75$ \\
        & \hspace{0.3em}- Empty-map Aug. & $78.23$ & $85.05$ \\
        \cmidrule(lr){2-4}
        & \texttt{LDGuid} U-Net & $\mathbf{82.83}^*$ & $\mathbf{88.68}^*$ \\
        \cmidrule(lr){2-4}
        & BIT (Base) & $84.04$ & $90.88$ \\
        & \texttt{LDGuid} BIT (NBR) & $\mathbf{86.63}^*$ & $\mathbf{92.54}^*$ \\
        \bottomrule
    \end{tabular}
    \vspace{-6mm}
\end{table}

Results on CaBuAr further depict the role of explicit semantic difference learning in more complex CD tasks, e.g., applications with multi-spectral data samples. As shown in Table \ref{tab:main_results}, \texttt{LDGuid} outperforms all baselines with IoU $86.63\%$ compared to the state-of-the-art $84.04\%$ achieved by BIT.

\paragraph*{Impact of Domain Knowledge} 
Our experiments with CaBuAr revealed that feeding raw 12-band Sentinel-2 data to DE can degrade performance (IoU $84.58\%$). This behavior can be explained by the fact that the high-dimensional data in this dataset contain significant noise, e.g., clouds and vegetation types. This can cause the DE module to quickly overfit. To overcome this issue, we pretrained DE with \textit{normalized burn ratios (NBRs)}, which is a specific index for combustion \cite{Firemon}. This forces the latent representation to focus on the burn signal. This demonstrates that for complex data, \texttt{LDGuid} can be efficiently deployed when the DE module is designed and pretrained with domain-specific knowledge.

\begin{figure}[t]
  \centering
  \includegraphics[width=\linewidth]{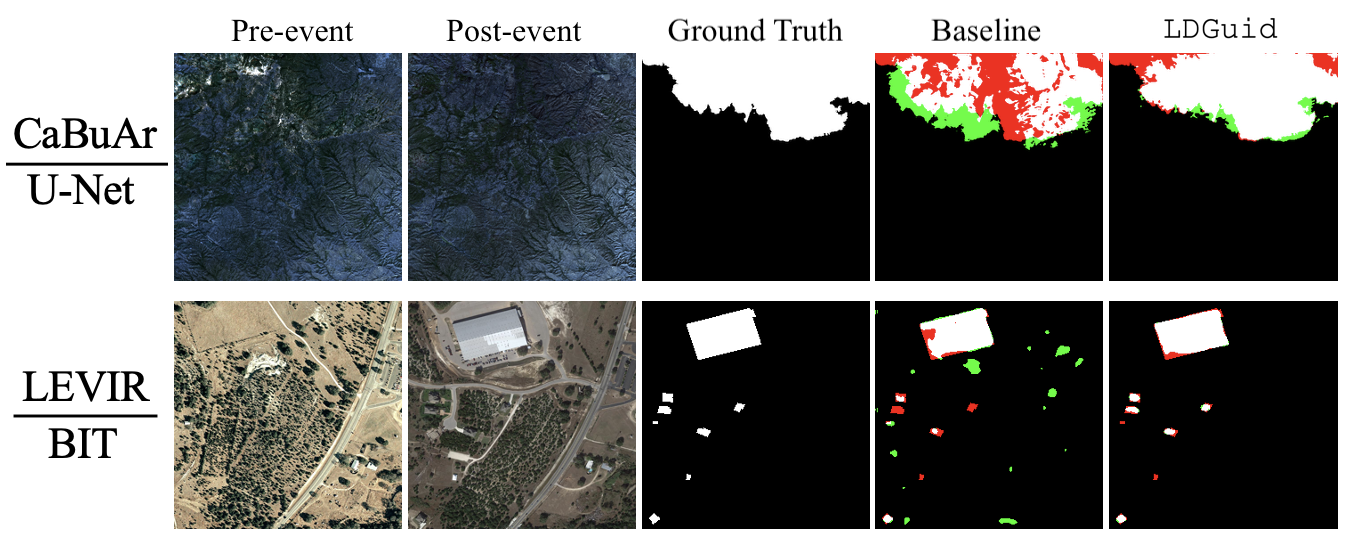} 
  \vspace{-6mm}\caption{Qualitative comparison on two distinct tasks. Top row shows results for CaBuAr using U-Net as the baseline. Bottom row shows results on LEVIR-CD with BIT as baseline. 
  In maps, white depicts correctly detected changes (true positive), while red and green highlight missed changes (false negatives) and false alarms (false positives), respectively. }
  \label{fig:multiple_images}
  \vspace{-5mm} 
\end{figure}

Fig.~\ref{fig:multiple_images} visualizes sample results from CaBuAr and LEVIR. As observed, \texttt{LDGuid} precisely delineates burn boundaries while the standard U-Net produces fragmented predictions. 

\paragraph*{Ablation Study}
To understand the behavior of the DE module, we investigate the sensitivity of the adversarial weight $\beta$ in the DE loss \eqref{eq:autoencoder_loss}. The results are reported for the SVCD dataset. As mentioned, $\beta$ balances the trade-off between the reconstruction fidelity and adversarial disentanglement. To observe this trade-off, we swept $\beta$ across a wide range from $0.1$ to $5$ and observed the DE pretraining performance.

As detailed in Table~\ref{tab:ablation_beta}, the results reveal two key insights:
\begin{inparaenum}
    \item The reconstruction loss remains stable across a broad range ($0.1 \le \beta \le 1.5$). This indicates that the DE pretraining is not overly sensitive to precise tuning.
    \item At small choices of $\beta$, DE fails to disentangle features, i.e., behaving like a standard AE. On the other hand, at excessive choices, the adversarial disentanglement constraint dominates the training and disrupts reconstruction. This illustrates the role of $\beta$ in balancing between the two objectives.
\end{inparaenum}
Throughout the experiments, we selected $\beta$ for each dataset to balance a fair trade-off. \vspace{-.25cm} 

\begin{table}[h]
    \centering
    \vspace{-2mm}
    \caption{\normalfont Impact of adversarial weight $\beta$ on DE pretraining on SVCD(Note: A higher Adv. Loss indicates better extracting of the latent difference).}
    \label{tab:ablation_beta}
    \setlength{\tabcolsep}{8pt} 
    \begin{tabular}{c|c|c} 
        \toprule
        \textbf{$\beta$} & \textbf{Rec. Loss} $(\downarrow)$ & \textbf{Adv. Loss} $(\uparrow)$ \\ 
        \midrule
        0.1    & $7.40 \times 10^{-6}$ & 0.330 \\
        \textbf{0.5} & \textbf{6.12 $\times$ 10\textsuperscript{-6}} & \textbf{0.314} \\ 
        1.5    & $6.14 \times 10^{-6}$ & 0.294 \\
        5.0    & $6.74 \times 10^{-5}$ & 0.311 \\
        \bottomrule
    \end{tabular}
    \vspace{-4mm}
\end{table}

\section{Conclusion}
\label{sec:conclusion}
We proposed \texttt{LDGuid}, a novel CD framework that explicitly learns semantic differences via a DE module. Invoking the information bottleneck method, we designed an efficient pretraining loop. 
We validated \texttt{LDGuid} across diverse datasets, including LEVIR-CD, WHU-CD, SVCD, and CaBuAr. The results demonstrate that \texttt{LDGuid} consistently outperforms the benchmark CD approaches, revealing the ability of \texttt{LDGuid} to capture domain-specific knowledge. Extending \texttt{LDGuid} to multi-modal CD tasks, e.g., Optical-SAR, is a natural direction for future work and is currently ongoing. 

\small
\bibliographystyle{IEEEtran}
\bibliography{references}

\end{document}